\documentclass{elsarticle}

\usepackage{lineno,hyperref}
\usepackage{booktabs}
\usepackage{amsmath}
\usepackage{algorithm}
\usepackage[english]{babel}
\usepackage[utf8]{inputenc}
\modulolinenumbers[5]

\usepackage{makecell}
\usepackage{comment}
\usepackage{adjustbox}
\usepackage{amssymb}
\usepackage{algpseudocode}
\algnewcommand{\LineComment}[1]{\State \(\triangleright\) #1}

\usepackage{todonotes}
\usepackage[version=4]{mhchem}
\usepackage{soul}
\usepackage{mathtools}

\usepackage{tcolorbox}
\usepackage{multirow}

\usepackage{algorithm}
\usepackage{algpseudocode}

\usepackage{amsmath,stackengine}

\stackMath

\journal{arXiv}

\begin{document}

\begin{frontmatter}
\title{ITI-IQA: a Toolbox for Heterogeneous Univariate and Multivariate Missing Data Imputation Quality Assessment}

\author[itiaddress]{Pedro Pons-Suñer}
\ead{pedropons@iti.es}
\author[itiaddress]{Laura Arnal}
\ead{larnal@iti.es}
\author[itiaddress,upvaddress]{J.Ramón Navarro-Cerdán}
\ead{jonacer@iti.es}
\author[itiaddress]{François Signol}
\ead{fsignol@iti.es}

\address[itiaddress]{ITI, Universitat Polit\`ecnica de Val\`encia, Camino de Vera, s/n, 46022 Val\`encia, Spain}
\address[upvaddress]{Universitat Politècnica de València, Camí de Vera, s/n. 46022 Val\`encia, Spain}

\begin{abstract}

Missing values are a major challenge in most data science projects working on real data. To avoid losing valuable information, imputation methods are used to fill in missing values with estimates, allowing the preservation of samples or variables that would otherwise be discarded. However, if the process is not well controlled, imputation can generate spurious values that introduce uncertainty and bias into the learning process. The abundance of univariate and multivariate imputation techniques, along with the complex trade-off between data reliability and preservation, makes it difficult to determine the best course of action to tackle missing values. In this work, we present ITI-IQA (Imputation Quality Assessment), a set of utilities designed to assess the reliability of various imputation methods, select the best imputer for any feature or group of features, and filter out features that do not meet quality criteria. Statistical tests are conducted to evaluate the suitability of every tested imputer, ensuring that no new biases are introduced during the imputation phase. The result is a trainable pipeline of filters and imputation methods that streamlines the process of dealing with missing data, supporting different data types: continuous, discrete, binary, and categorical. The toolbox also includes a suite of diagnosing methods and graphical tools to check measurements and results during and after handling missing data.

\end{abstract}


\end{frontmatter}

\newpage

\section{Introduction}
\label{sec:intro}

Missing data is a common problem in many fields surrounding data processing and analysis. Missing values in a dataset can arise from various sources, such as errors during data entry, individuals' unwillingness to answer some questions, invalid data types, intentional omissions or by design, equipment malfunctions, etc. Ignoring or tackling this problem incorrectly often leads to the loss of information, poor performance of machine learning models, and biased analysis and predictions, among other potential issues. Therefore, acknowledging and carefully dealing with this problem is critical for any data science project to succeed.

Dealing with the issue of missing data is a complex task, influenced by numerous factors such as the volume of missing data, the nature of the missingness, the data type, statistical assumptions, cost constraints, and data quality requirements for a specific task. Despite the array of potential solutions, only a few are suitable for a given task, and finding the best path of action can be challenging and time-consuming.

Some modern machine learning models are prepared to deal with datasets containing missing values. For example, tree-based algorithms like CART \cite{breiman2017classification} or XGBoost \cite{chen2015xgboost} can naturally handle missing values, ignoring them when determining the best split or treating them as separate categories. Using these models to ignore missing values helps to simplify the data processing steps. Nevertheless, it limits the use of machine learning to those that support missing data. Besides, this can lead to ignoring potentially valuable information hidden behind missing data that could be leveraged otherwise. If there is any identifiable pattern in data missingness, this can also introduce potential biases in the model that could, in turn, negatively impact performance or produce unreliable results and misleading conclusions.

Some modern machine learning models are designed to handle datasets with missing values. For instance, tree-based algorithms like CART \cite{breiman2017classification} or XGBoost \cite{chen2015xgboost} can naturally handle missing values, either by ignoring them when determining the best split or by treating them as separate categories. While this simplifies data processing, it restricts the use of machine learning solutions to those that support missing data. Moreover, it can lead to neglecting potentially valuable information hidden behind missing data that could be leveraged otherwise. If there is any discernible pattern in data missingness, this can introduce biases in the model that could compromise its performance or produce unreliable results and misleading conclusions.

Another and often easiest solution is removing missing values from the dataset. This can be achieved in two ways. One consists of eliminating all samples that contain any missing value. The second one involves erasing features where missing values are present. While simple, these approaches must be carefully considered. If a dataset has missing values randomly distributed across the dataset, these methods can lead to an enormous loss of information, even disqualifying a dataset to be useful for any study. An intermediate solution is to set specific data quality criteria for samples and features, discarding only instances that do not meet a given completeness threshold. However, this still leaves missing values in the dataset that should be handled differently. Conversely, higher quality thresholds result in more substantial information loss, which can lead to even poorer performances. This tradeoff between quality standards and information loss has to be carefully considered and often lacks an ideal, clear answer.

The last solution, but not less important, is to use data imputation methods to fill in missing values. Univariate methods fill missing values in a given variable based on information contained in other observations of that concrete variable. Alternatively, multivariate methods do it by leveraging the information in other variables of the same observation, learning from the relations observed in other individuals with complete entries. Univariate methods are based on statistical assumptions that could be false for a given feature. Multivariate methods may produce better results when there are complex relations in the data, but often demand high computational costs. Although not free of disadvantages, imputation methods that learn from the existing data, be it transversally or longitudinally, are usually the most effective way of addressing missing data.

Despite having presented some solutions to the problem separately, the best path of action often consists of a combination of methods. For example, basic quality filters can discard samples and features with excessive missing values for the remaining ones to be filled in with imputation models. If any feature cannot be correctly imputed, a machine-learning model resilient to missing values can be utilized. A notion worth mentioning is that when no imputation method seems reliable enough, the best way to deal with missing values could be to leave them as such and use a prediction model that supports them, if not removing samples and features containing them altogether. In \cite{schafer2002missing}, authors state that \textit{ad hoc} amendments for reducing missingness can do more harm than good, and researchers should study the nature of missingness itself before deciding on its treatment. In \cite{janssen2010missing}, authors advise that using simple imputers can lead to misleading results, and advocate for not using these in favour of multiple imputation, remembering that the purpose of this task is to prevent the exclusion of observed data, not to create new values.

In this work, we propose a tool called ITI-IQA to assist in the management of missing data in any tabular dataset. It can be used to determine the best imputer for each variable by combining the quality of its imputation and completeness in a single objective score. Besides, additional statistical tests are proposed to prevent the inclusion of biases that the non-random nature of missingness could bring.

\section{Materials and methods}

\label{sec:matnmet}

\subsection{Software}
\label{sec:software}

In the following experiments, we have used Python 3.10.12 together with widely used libraries for data preprocessing, analysis, visualization and machine learning. The mainly used libraries, along with their specific versions and purposes, are listed in Table \ref{tab:libraries}.

\begin{table}[h]
\centering
\caption{Main Python libraries used in this work.}
\label{tab:libraries}
\begin{tabular}{lll}
\toprule
\textbf{Name} & \textbf{Version} & \textbf{Purpose} \\
\midrule
Numpy & 1.23.5 & General math computing\\
Pandas & 1.1.5 & Data structuring, preprocessing\\
Scipy & 1.13.0 & Data analysis\\
Matplotlib & 3.5.3 & Visualization\\
Missingno & 0.5.1 & Visualization (missing data)\\
Scikit-learn & 1.5.0 & Machine learning, imputation\\
XGBoost & 2.0.3 & Machine learning (classification)\\
\bottomrule
\end{tabular}
\end{table}

\subsection{Main algorithm design}
\label{sec:matnmet_alg}

For all the reasons mentioned before, imputation is not a trivial process. It is essential to guarantee a minimum level of quality when doing any imputation, which means measuring this quality is necessary. If the quality is insufficient, either a predictive model that tolerates missing values must be chosen, or the variables must be excluded from the study.

The quality of imputation can be measured on the basis of existing values (groundtruth) by considering them as temporarily missing, imputing them, and measuring the similarity between the imputation and the true value. The IQA's main procedure for assessing the quality of variables is based on this central idea. IQA integrates feature completeness and its imputation quality into a single quality metric. The reasoning behind this is that if a variable is fully observed, then no imputation is needed; conversely, if a variable could be imputed perfectly, its completeness does not matter.

The remainder of this section thoroughly details the steps of this procedure. The IQA main algorithm takes four items as input:
\begin{itemize}
\item A dataset $D$ with a set of variables $X$ whose quality will be assessed.
\item An splitter $S$ that will divide the data into train and evaluation sets.
\item A set of missing-values imputer methods $\{I_1, I_2, \cdots, I_n\}$.
\item An scorer function $\lambda$ to assess the similarity between observed and imputed values.
\end{itemize}

Other parameters to consider are:
\begin{itemize}
	\item Quality threshold $\tau$: once a final quality score is computed for a variable, if $\tau$ is supplied and the score does not surpass it, the variable will be discarded at the end of the procedure.
	\item An encoder to convert variables of categorical type to numerical variables. This encoder can be chosen to process variables automatically detected as categorical, since IQA cannot handle these in their original form as of the date of this publication. More on this limitation of the current state of the tool can be found in Section \ref{sec:future}.
\end{itemize}

The following item list describes the main steps in the procedure to get a final quality score for each variable. Keep in mind that the items in this sequence are presented in a more comprehensive order that could not match the actual arrangement, which is actually designed with parallelization and resource optimization in mind.

\begin{itemize}
	\item For each variable $x \in X$, and for each imputer $I_i$, get its imputation score $\delta^i_x$:
	\begin{itemize}
		\item A copy of the dataset $D$ is made, $D^{\prime}$.
		\item Splits are generated using the splitter $S$ to create separate train and evaluation sets. For each split:
		\begin{itemize}
		\item $I_i$ is trained with the train set to learn how to impute $x$.
		\item We synthetically generate missing values in $D^{\prime}[x]$ by treating existing complete values as temporarily missing. As the original values are preserved in $D$, this will allow us to measure the difference between the imputed values and the actual values.		
		\item The fitted $I_i$ is used to fill in the newly created missing values in $D^{\prime}$.
		\item The scorer function $\lambda$ is used to measure the similarity between the original $x$ in $D$ and the imputed $x$ in $D^{\prime}$. This score for the imputer $i$ and the feature $x$, $\delta^i_x$, is saved.
		\end{itemize}
	\end{itemize}
	\item For each variable $x \in X$, the imputer yielding the best score is selected. The score obtained using this imputer becomes its imputation score $\delta_x$.
	\item For each variable $x \in X$, get its fraction of completeness $\mu_x$ (number of observed values divided by the total number of entries, missing and observed).
	\item For each variable $x \in X$, get its final quality score $\omega_x$ as in Eq. \ref{eq:qs}:
		\begin{equation}
		\omega_x = \mu_x + (1 - \mu_x) * \delta_x
		\label{eq:qs}
		\end{equation}
		
\end{itemize}

Since $\mu_x$ and $\delta_x$ are both values ranging between 0 and 1, $\omega_x$ is a score also ranging from 0 to 1. This way, on the one hand, if a variable is 100\% complete, $\mu_x=1$ so that, whatever the imputation score $\delta_x$, the final quality will be $\omega_x=1$. If, on the other hand, a variable has high missingness, but an imputer can estimate values with perfect measured performance ($\delta_x=1$), then it would also be true that $\omega_x=1$. Again, the reasoning is that if a variable is fully observed, imputation is unnecessary. On the other hand, if a variable can be perfectly imputed, its completeness is irrelevant.

If a minimum quality threshold $\tau$ is provided, variables can be discarded if their quality score ($\omega$) is lower than $\tau$. Variables that do not meet the quality criterion $\tau$ are removed at the end of the whole process so that they can still participate in the imputation of the other features.

If more technical detail is preferred, Algorithm \ref{alg:missing_values} details, with a simplified pseudocode, the procedure to get a weighted quality score for each feature in the dataset, and filtering features based on a chosen quality threshold.

\begin{algorithm}
\small
    \caption{Column filtering based on missing values and imputation quality}
    \label{alg:missing_values}
    \begin{algorithmic}[1]
    
    \Procedure{get\_imputation\_score}{$data$, $imputer$, $splitter$, $scorer$}
    \LineComment{Create a data copy and replace complete values with imputer predictions}
    \vspace{3pt}
    \State $data\_imputed \gets data$ 
    
    \For{$col$ in $data$.columns}
        \For{$train\_index$, $test\_index$ in splitter($data$)}
        	\State $train\_set \gets data[train\_index]$
        	\State $test\_set \gets data$[$test\_index$]
        	\vspace{3pt}
    		\State $imputer$.fit($train\_set$) \Comment{Train imputer with train data}
    		\vspace{3pt}
    		\State $test\_set[col]\gets $ MissingValue \Comment{Replace real value with missing}
    	\State $data\_pred \gets imputer$.transform($data\_test$) \Comment{Fill in missing values}
    	\State $data\_imputed[col] \gets data\_pred[col]$ \Comment{Store prediction}
    
    \EndFor
    \EndFor
    \\
    \LineComment{For each column, get score of original vs. predicted values}
    \vspace{3pt}
    \State $imputation\_score \gets$ dictionary()
    \For{$col$ \textbf{in} $data$.columns}
    	\State $col\_score \gets$ scorer($data[col]$, $data\_imputed[col]$)
    	\State $imputation\_score$[$col$] $ \gets col\_score$
    \EndFor
    \State \textbf{return} $imputation\_score$
    \EndProcedure
    \\
    
     \Procedure{get\_missing\_fraction}{$data$}
	\State $missing\_fraction \gets$ dictionary()
	\State $num\_rows \gets$ get\_number\_rows($data$)
    \For{$col$ \textbf{in} $data$.columns}
    	\State $col\_nan\_fraction \gets$ sum($data[col] ==$ MissingValue) / $num\_rows$
    	\State $missing\_fraction$[$col$] $\gets col\_nan\_fraction$
    \EndFor
	\State \textbf{return} $missing\_fraction$
    \EndProcedure
    \\
    
    \Procedure{filter\_missing\_quality}{$data$, $columns$, $imputer$, $splitter$, $scorer$, $threshold$}
    
    \LineComment{Get a score weighting missing value counts and imputation score}
    \vspace{3pt}
    
	\State $imputation\_scores \gets$ get\_imputation\_score($data$[$columns$], $imputer$, $splitter$, $scorer$)
	\State $missing\_fractions \gets$ get\_missing\_fraction($data$[$columns$])
    
    \For{$col$ \textbf{in} $columns$}
    \State $imp\_score \gets imputation\_scores$[$col$]
    \State $mis\_fraction \gets missing\_fractions$[$col$]
    \LineComment{Considering that the maximum score equals 1:}
    	\State $quality\_score \gets mis\_fraction \cdot imp\_score + (1 - mis\_fraction)\cdot 1$
    	\vspace{3pt}
    	\If{$quality\_score < threshold$}
    	\State $data \gets $drop\_column($data$, $col$) \Comment{Drop cols with score $<$ threshold}
    	\EndIf
    \EndFor
    \State \textbf{return} $data$
    \EndProcedure

    \end{algorithmic}
\end{algorithm}

\subsection{Bias check and correction}
\subsubsection{Statistical tests to discard imputations out of distribution}
\label{sec:statchecks}

When dealing with missing data, it is important to know about the different missingness mechanisms described in the bibliography. ``Missing Completely At Random'' (MCAR) mechanism corresponds to a distribution of missingness that does not depend on observed data nor unobserved data, as it appears to have emerged from a truly random sampling process \cite{graham2009missing, schafer2002missing}. This is a convenient property that, realistically, usually fails to hold. In ``Missing At Random'' (MAR), the missingness does depend on observed data but not on missing data. Depending on observed data means that the probability of a value being missing in an attribute can be explained by the observed data in the other attributes. This is a more general property that imputation methods often assume. However, there is a third mechanism. When the previous conditions are false, and the missingness distribution depends on missing data, we talk about ``Missing Not At Random'' (MNAR). In MNAR, missingness is considered to be ``informative'' because it is related to the missing values themselves, even after considering the observed data. Since the probability of missing values depends on unobserved data, then the missingness does hold some sort of information that could be valuable. The problem is that most imputation techniques assume non-MNAR properties, and failing this assumption can lead to the introduction of dangerous biases. With the intent of alleviating this problem, statistical tests have been implemented to detect biases in imputed data.

Imagine, for instance, a binary classification task where an attribute has a higher chance of being missing when the class is positive. Suppose a simple imputer replaces missing values with a constant (say, the average of the attribute) that is not necessarily commonly observed in the actual data. In that case, all instances with missing values now have this same constant and distinguishable value. Any simple predictive model could then recognize a pattern where this constant is commonly associated with the positive class, indirectly associating previous missingness with the target class. This way, an attribute can be picked as relevant for making a prediction when the truly useful information is contained in missingness instead of the observed values. This scenario can result in unexpected model behaviour in real use cases, where the high performance observed during training and evaluation may not be reproducible.

In order to avoid the introduction of biases, statistical tests are performed to assess if the imputed data could come from the same distribution as the observed data. If the null hypothesis (they come from the same distribution) can be rejected with sufficient statistical significance, then any model could seize this bias in a non-intended way. Since different tests are suited for different data types, we must first decide on a criterion to separate continuous and other types of attributes. Here, a minimum appearance frequency of 5 of every unique value is required for a variable to be considered discrete or nominal, as recommended in \cite{yates2002}. The Kolmogorov-Smirnov test is applied to continuous variables, while the Chi-square test of independence is applied to the remaining ones. If a variable's imputed values are found to be differentiable from the original ones, then the imputer method producing those values is discarded for the imputation of said variable. 

\subsubsection{A Priori Probabilistic Random Imputer}

A new univariate imputer that preserves the same distribution as the original data has been developed for this study. This imputer, called the ``a priori probabilistic random'' (APPRandom) imputer, has the same effect as drawing random samples from the non-missing values of a feature to fill missing values in said feature, which is different than assigning random values from a range of observed values. This way, unlike some other imputers that can introduce new values never observed in the complete data, APPRandom always fills missing values with already observed values. Moreover, the random sampling method preserves the probability distribution of the observed values. This design could help to avoid introducing biases that other simple imputers could create since imputed values are not statistically distinguishable from the original.

When no imputation methods are found to be suitable for a given variable, for example, after applying the checks described in Section \ref{sec:statchecks}, then the APPRandom method is used, as it is designed to minimise the risk of introducing bias in the imputed data.

\subsubsection{Pseudo-rounding of non-normal data}

When imputing binary, categorical, or even discrete data, imputers drawing predictions from a normal distribution are prone to result in implausible estimations. This is also true for simple imputers. For example, a mean imputer is very likely to fill in missing values in binary data (0 --- 1) with a value that is not 0 nor 1, but a value in between that is never observed in reality. This would constitute an immediately recognisable bias, and may lead to poor performance and flawed analysis. In these cases, many researchers resort to rounding (censoring) disallowed values to the closest observed value so that the original distribution is better preserved, but this can also result in the introduction of biases, especially in binary features, as described in \cite{horton2003potential}.

In their work, \cite{bernaards2007robustness} proposed two alternative methods to the simple rounding technique for binary features: ``coin flipping'' and ``adaptive rounding'' techniques, the latter providing the best performance and producing lower bias according to their findings. Adaptive rounding is similar to simple rounding, where each value is rounded to either 0 or 1, with the cut-off point being a threshold based on a normal approximation to the binomial distribution. Many other pseudo-rounding methods have been presented, but no single method has emerged as the definitive one \cite{van2018flexible}. Because of its simplicity and given its slight edge over other techniques, in this work, the adaptive rounding method has been implemented to pseudo-round binary features. Additionally, to round other discrete, non-binary features, a censoring operation to the closest observed value has been applied.

\subsection{Dependency graphs for multivariate imputation optimization}
\label{sec:graphs}

Despite their powerful capabilities, one of the main disadvantages of multivariate imputers is their high computational costs, especially when applied over big datasets with a high amount of variables, when very complex relations exist between them, or when many variables have missing values simultaneously. In such cases, the imputers can struggle to handle the missing data patterns and may require much longer processing times and a significant increase in computational resources needed. In this work, we try to reduce the impact of this disadvantage by using custom dependency graphs for each dataset.

If a variable \textit{A} is redundant with \textit{B}, so that any of them could be accurately estimated by watching the other, then we note \textit{B}$\leftrightarrow$\textit{A}. These dependencies often have one single direction. For example, imagine the variable \textit{A} being Obesity, defined as a Body Mass Index (BMI, calculated from weight and height) over 30, and a second variable \textit{B} being the BMI itself. Then, \textit{A} can be predicted with perfect accuracy if \textit{B} is known, but \textit{B} can only be poorly approximated if the known variable is \textit{A}. In this case, we say that \textit{B}$\rightarrow$\textit{A}. Unless a dataset has already been processed to remove any redundancy or correlation, finding these kinds of relations is widespread, as many variables have some shared information. In some cases, relations linking many variables can be found. For example, if \textit{A}=Obesity, \textit{B}=BMI, and \textit{C}=Weight, we could imagine that \textit{A}$\leftarrow$\textit{B}$\leftrightarrow$\textit{C}. Then, knowing the value of \textit{C}, it could be possible to fill in missing values in \textit{A} and \textit{B} more confidently than using a random imputer. In this section, we describe our efforts to take profit from these dependency graphs in order to empower the multivariate imputation process.

The knowledge of the dependency graph of a given variable could also be beneficial when specific missingness patterns such as file-matching, multivariate, or monotonic (as described in \cite{rubin2018multiple}) exist in a dataset. In a file-matching scenario, i.e., if a variable \textit{A} is missing when \textit{B} is not, and vice versa, then the imputer cannot learn to infer one directly based on the other since they never appear together. A similar situation applies if a multivariate missing pattern involves both \textit{A} and \textit{B} (they are always missing together), as they may be observed together in training but not at imputation. This could apply to monotone patterns to some extent, too. As it is not likely that \textit{A} and \textit{B} will appear in graphs together, the algorithm will never consider imputing one from another, potentially saving valuable time and resources.

For any dataset with a set of variables $X$, the dependency graph for each single variable $x$ is constructed following these steps:
\begin{itemize}
\item $x$ is extracted from $X$, so that the remaining dataset contains $X \setminus x$. In this scenario, $X \setminus x$ will constitute our new set of predictive features, and $x$ will become our new target feature.
\item We train a predictive model (in a sense, regressor models can handle a broader set of target data types) that receives $X \setminus x$ as input and tries to predict $x$. Then, using a reserved test set, we measure the model performance $S_{X \setminus x}$, for example, using $R^2$ score.
\item Iterating through the variables in $X \setminus x$, we permute each variable in the test set and measure the model performance again, obtaining $S^\prime_i,~i \in X \setminus x$. Subtracting $S_{X \setminus x} - S^\prime_i$ gives us the permutation importance of the variable $i$. The greater the contribution of $i$ for predicting $x$ is, the lower $S^\prime_i$ will be, thus greater the subtraction.
\item Variables with the greatest permutation importance are those most predictive towards $x$. Therefore, if one wanted to impute $x$, the priority features to consider for training an imputer would be those with greater importance. The set of $N$ most informative features can be chosen, limiting $N$ to a desired maximum number, establishing an importance threshold, or a combination of both restrictions. If this process is applied to every feature in $X$ (so that each one becomes our target $x$ at some point), each of the top $N$ informative variables to predict $x$ may have its own set of top informative features. This way, a graph of dependencies is constructed.
\end{itemize}

\begin{figure}[H]
\begin{center}
\includegraphics[width=0.2\textwidth]{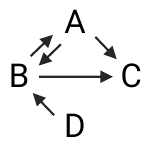}
\end{center}
\caption{Dependency graph containing four features: A, B, C, D. An arrow from A to B indicates that B is dependent on A, or that A is important for predicting B.}
\label{fig:graph}
\end{figure}

The natural tendency we have observed when performing this process is for variables to form clusters of related measurements. We can use the graph in Figure \ref{fig:graph} as an example. Here, in graph containing $\{A, B, C, D\}$, a variable $A$ depends on $B$, $B$ depends on $A$ and $D$, and $C$ depends on $B$ and $A$. Then, for imputing $A$, we should train an imputer with $\{B, D\}$ plus $A$ itself. While it is true that only the variable $B$ is directly pointing at $A$, in the case that $B$ is also missing, an imputer that is also able to see $D$ would be more robust since it could be able to estimate the missing link $B$ when $D$ is complete. Additionally, feature $D$, which has no inward edges, could only be imputed by a univariate imputer that only sees the complete values in $D$.

In the case that a dependency graph is calculated for a dataset, it should be provided to the IQA algorithm as a dictionary, where every key represents a feature, and its value is the list of features on which it depends. Following the example in Figure \ref{fig:graph}, and extending the dependency list of each feature as much as the graph allows, this dictionary would take the form:
\begin{verbatim}
{
    A: [B, D],
    B: [A, D],
    C: [B, A, D]
}
\end{verbatim}
If one chooses to use dependency graphs in IQA, its algorithm must be slightly adjusted. If we name $\Delta$ to the dictionary built from a dependency graph, in Algorithm \ref{alg:missing_values}, $\Delta$ would be passed as an input in procedure GET\_IMPUTATION\_SCORE. Then, in lines 6 and 7, the train and test sets for a given feature \textit{col} would be constructed using only the features in $\{\Delta[col], col\}$. Following the running example, an imputer trained to impute $A$ would be able to see features $\{B, D, A\}$.

\subsection{Toolbox}

\label{sec:utilities}

The tool implementing the IQA main algorithm has been designed with easy usability and configuration flexibility in mind. For this end, the methods are configurable with structured configuration files that are intuitive to fill while enabling splitters, scorers, and imputers of any source and nature, provided that they implement some mandatory methods:

\begin{itemize}
\item Splitters (train-test splitters) must be classes implementing a \textit{split} method and returning train and test indexes, plus groups optionally.
\item Scorers (or methods to get error metrics) must be functions that take true and predicted arrays, may accept other scorer params, and return a numeric result.
\item Predictors (imputer methods) must be instances of imputer classes that implement the usual \textit{fit} method and either a \textit{transform} or \textit{predict} method.
\end{itemize}

\begin{figure}[H]
\centering
\includegraphics[width=0.6\textwidth]{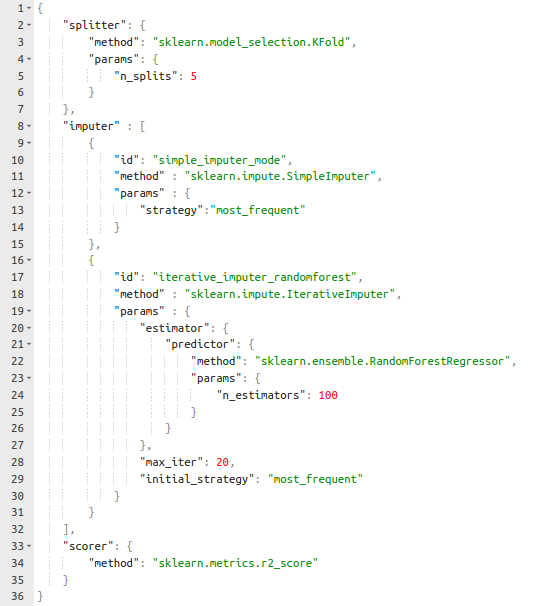}
\caption{Example of configuration JSON file for IQA.}
\label{fig:iqaconfig}
\end{figure}

All classes, instances, or methods can also be introduced as a \textit{module.import} string in a configurable JSON file. Figure \ref{fig:iqaconfig} shows an example of this file.

The presented toolbox accommodates various utilities to review the algorithm results and assess final data quality. Plotting functions allow users to visually assess data quality, check the compatibility of distributions between observed and imputed data, and better understand the algorithm's results.

\begin{figure}[H]
\centering
\includegraphics[width=0.75\textwidth]{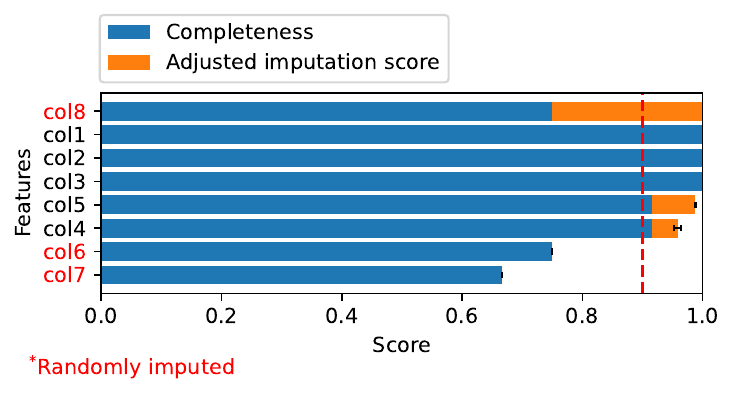}
\caption{Example of IQA final quality results. A quality threshold of 0.9 has been set so that \textit{col6} and \textit{col7} would be removed while, in principle, the rest is accepted. \textit{col8}, which is a constant variable, can be perfectly imputed despite having been asigned an APPRandom imputer, which is indicated by its red color.}
\label{fig:iqaresults}
\end{figure}

Regarding plotting capabilities, Figure \ref{fig:iqaresults} shows the main graphical representation, which ranks the features according to their quality score after performing the missing data processing steps proposed by IQA. For each studied feature, a blue bar shows its completeness (fraction of non-missing values), while the orange portion represents the imputation score adjusted to the missing fraction. The standard deviation of this score across splits is also shown. The sum of both bars is equal to the quality score $\omega$ (Equation \ref{eq:qs}), between 0 and 1, presented in Section \ref{sec:matnmet_alg}. If chosen, the quality threshold to accept or not each feature is represented as a dashed line. Moreover, features that require the use of APPRandom imputer to fill in their missing values have their names appearing in red so that the user can decide whether to include these variables.

Additionally, a function for recommending a number of iterations for multiple imputation has been implemented. In \cite{little1987multiple}, the author described that the efficiency ($\epsilon$) of an estimate is approximately

\begin{equation}
\epsilon = \left( 1 + \frac{\gamma}{m}\right)^{-1},
\label{eq:efficiency}
\end{equation}

where $m$ is the number of imputations and $\gamma$ is the fraction of missing information in the variable being imputed. Reordering this equation, the number of imputations for a given variable with an approximately desired efficiency can be calculated as

\begin{equation}
m = \gamma \left(\frac{1}{\epsilon} - 1 \right)^{-1},
\label{eq:numberm}
\end{equation}

Equation \ref{eq:efficiency} proves that efficiency gains rapidly diminishes after the first few rounds of imputations. For example, for a variable with 50\% missing values, 10 imputations would achieve ~95\% efficiency, while 20 imputations, double the initial amount, would get ~98\% efficiency. This marginal increase could or could not seem worth doubling the efforts and resources taken, depending on the user's needs and the cost of each additional imputation.

\subsection{Evaluation and comparison with other imputation solutions}
\subsubsection{Evaluation methods}
\label{sec:evalmethods}

For IQA to function in the following experiments, we must choose a collection of imputers. Regarding univariate (simple) imputation strategies, we use mean, median, and mode imputation \cite{SimpleImputer}. These replace missing values of a feature with the mean, median, and most frequent values of the non-missing values of said features, respectively. The mean and median strategies can create non-existing values in the original data. The newly introduced APPRandom imputer, which falls in the univariate imputers category, is also added to the set of imputers.

Additionally, we will consider the following multivariate imputation techniques:

\begin{itemize}
\item N-Nearest Neighbour imputation \cite{troyanskaya2001missing}, replaces each missing value in a feature by the mean value of the values in said feature of the N nearest instances based on the values of the non-missing values in other features.

\item Iterative imputation \cite{Buuren2011, Buck1960}, starts from a simple imputation strategy (mean, mode, etc.) and iterates using a more complex predictive model to estimate the replacement of a missing value based on observations of non-missing entries. The method is comparable to the popular statistical method MICE (standing for Multivariate Imputation by Chained Equations) \cite{azur2011multiple}. Making multiple imputations, as opposed to single imputations, accounts for the statistical uncertainty in the imputations. Machine-learning estimators powering this technique will be Bayesian Ridge Regression \cite{tipping2001sparse}, and Random Forest Regression \cite{liaw2002classification}.
\end{itemize}

Table \ref{tab:imputers} shows the collection of univariate and multivariate imputers used for evaluating IQA. The number of estimations for iterative imputers ($m$) has been set to 20, which, following the formulas presented in Section \ref{sec:utilities}, would aim for 95\% efficiency even in the most extreme cases where approximately 99\% of values are missing. Although it is very unlikely that we will encounter this situation, we choose this conservative value to remove uncertainty from successive steps due to using too few imputations. In \cite{schafer2002missing}, authors claim that, even if $m$=10 is enough to achieve 95\% efficiency with 50\% of missing values, researchers usually prefer to remove noise from other statistical summaries and, in many practical scenarios, $m$=20 effectively achieves this. Moreover, authors in \cite{graham2007many} suggest that researchers should use more imputations than previously recommended.

\begin{table}[h]
\caption{Imputers used for evaluating IQA.}
\label{tab:imputers}
\centering

\begin{tabular}{ccll}
\toprule
\textbf{Imputer} & \textbf{Type} & \textbf{ID} & \textbf{Parameters}\\
\midrule
\multirow{1}{*}{SimpleImputer}
	& Univ. 
	& mean 
	& strategy: mean\\
    \hline
\multirow{1}{*}{SimpleImputer}
	& Univ. 
	& median 
	& strategy: median\\
    \hline
\multirow{1}{*}{SimpleImputer}
	& Univ. 
	& mode
	& strategy: mode\\
    \hline
\multirow{1}{*}{APPRandom}
	& Univ. 
	& random 
	& ---\\
    \hline
\multirow{1}{*}{NNeighbors}
	& Multiv. 
	& knn3 
	& n\_neighbors: 3\\
    \hline
\multirow{1}{*}{NNeighbors}
	& Multiv. 
	& knn5
	& n\_neighbors: 5\\
    \hline
\multirow{1}{*}{NNeighbors}
	& Multiv. 
	& knn10
	& n\_neighbors: 10\\
    \hline
\multirow{5}{*}{IterativeImputer}
	& \multirow{5}{*}{Multiv.} 
	& \multirow{5}{*}{iter\_br} 
	& init\_strategy: mode\\
	&&& max\_iter: 20\\
	&&& estimator: BayesianRidge(\\
	&&& $\hookrightarrow$ max\_iter: 300\\
	&&& )\\
    \hline
\multirow{5}{*}{IterativeImputer}
	& \multirow{5}{*}{Multiv.} 
	& \multirow{5}{*}{iter\_rf} 
	& init\_strategy: mode\\
	&&& max\_iter: 20\\
	&&& estimator: RFRegressor(\\
	&&& $\hookrightarrow$ n\_estimators: 100\\
	&&& )\\
    \hline
\multirow{5}{*}{IterativeImputer}
	& \multirow{5}{*}{Multiv.} 
	& \multirow{5}{*}{iter\_xgb} 
	& init\_strategy: mode\\
	&&& max\_iter: 20\\
	&&& estimator: XGBRegressor(\\
	&&& $\hookrightarrow$ n\_estimators: 100\\
	&&& $\hookrightarrow$ max\_depth: 6\\
	&&& $\hookrightarrow$ learning\_rate: 0.1\\
	&&& )\\
    \hline

\end{tabular}
\end{table}

In order to test the IQA tools, we will use a public dataset from UCI Machine Learning Repository \cite{kelly2023uci}, an open data repository well-established in the machine-learning field. This dataset is described in Section \ref{sec:dataset}. If necessary, new missing values will be artificially introduced in a randomized way, following no intended pattern (values completely missing at random).

The first test is designed to measure the capability of IQA of preprocessing data with missing values so that a machine learning model trained with this data can solve a specific task with decent accuracy, compared to imputations performed by other means.

First, the IQA algorithm will be used to estimate the best way to handle missing data. Then, in a K-Fold experiment, the retrieved preprocessing pipeline will be applied, producing a dataset with no missing values. The imputers will always be trained with observations in the training sets and then used to transform (fill missing values) observations in both train and test sets. Otherwise, it could result in over-optimistic performances. Using the preprocessed tranining data and a target array, an XGBoost model will be trained to solve a specific classification task. Then, test data will be used to evaluate the model performance. The achieved performances will be compared to those obtained using other missing data handling techniques.

Concretely, model performances, measured in AUROC, will be measured after preprocessing data at different levels of general missingness: 25\%, 50\%, and 75\% missing values.

Aside from model performance, the quality of the imputed data will be assessed. In the past, some authors \cite{jager2021benchmark, schafer2002missing} have used error metrics like RMSE to benchmark and compare imputation methods, and metrics like F1-Score to assess the quality of categorical data imputation, more akin to classification tasks. However, other researchers \cite{casiraghi2023method, rubin2018multiple} have stated that using error metrics, such as MAE or RMSE, to compare original and imputed values is a flawed technique since it does not account for the many ways in which imputation can do more harm than good, mainly through the inclusion of biases in estimates.

In this work, the similarity between original and re-imputed values in continuous and discrete numerical features is measured using any scoring function that can work with continuous data, such as $R^2$, although the final scoring function will depend on the overall nature of the data. Meanwhile, binary categorical data is assessed using balanced accuracy. While simple accuracy would not work well with quasi-constant features (yielding a high score even when the imputed value is always the same as the majority value), balanced accuracy is well prepared for these imbalanced scenarios as it is the arithmetic mean of sensitivity and specificity. While these scoring functions will be used to calculate imputation scores in IQA and then select the optimal imputation steps, for the reasons mentioned above, they will not be used for comparing the goodness of imputation of different imputation strategies. Instead, the ability of imputers to fill in missing values in an ``undetectable'' manner will be measured. As stated before, ``imputation is no prediction'': the objective is not to minimize the total error between observed and estimated values, but not to compromise subsequent analysis and conclusions. When imputed values differ significantly from observed values in a recognisable pattern, imputed values could be easily identified by a capable machine learning model, therefore learning which samples contained missing values initially. This can potentially introduce dangerous biases during the training phase or even leak information about the target we are unaware of. To assess this, the following strategy will be followed for each missingness fraction (original, 25\%, 50\%, 75\%):
\begin{itemize}
\item The dataset $D$, containing a set of variables $X$, is successively split into train and test sets using a 5-Fold strategy. In each fold, the chosen imputers are trained with the train set and used to fill in missing values in the test set. Then, the five transformed test-set slices of the dataset are concatenated to build $D^\prime$, a copy of the same size and shape as $D$ but with no missing values.
\item Cycling through each feature $x \in X$, a new binary classification target $y_x$ is constructed, where a positive value indicates that the entry at that position in $x$ is missing and therefore has been imputed in $D^\prime$.
\item Following the same 5-Fold strategy as the first step, $D^\prime$ and $y_x$ are divided into train and test data and target, respectively. The training slice of $D^\prime$ and $y_x$ are used to train a classification model that tries to predict $y_x$ watching the features in $D^\prime$. Then, the predictions using the reserved test set are compared with the test entries of $y_x$, and the model's AUROC performance score is calculated.
\item Mean and 95\% confidence interval AUROC for each feature $x$ participating as $y_x$ are calculated. If this score is high for a feature $x$, then it means that a model could recognise if observations in this feature are originally observed or imputed.
\end{itemize}

Lastly, an alternative candidate imputation method is selected to compare and contrast our results. MICE is a particular multiple imputation technique \cite{van2011mice}. First, a simple imputation technique (e.g., mean or mode) imputes missing values in each variable, and then an imputer performs several rounds of imputations. In each iteration, each variable is re-imputed by a separate model exploiting the values precedingly imputed in the other variables, thus ``chaining'' successive imputations \cite{casiraghi2023method}. Several studies consider multiple imputation as the most flexible valid missing data approach among those that are commonly used \cite{li2015multiple}, with MICE being the most popular implementation \cite{jager2021benchmark}. For this reason, we will compare our results with those obtained using scikit-learn \cite{scikit-learn} for implementing MICE and using this tool to impute all missing values. Moreover, for its ease of use and overall popularity, K-Nearest Neighbor Imputer \cite{10.1093/bioinformatics/17.6.520} will also be used in the comparison of the results. Lastly, a pipeline consisting only of our APPRandom imputer will also be used to serve as a more naïve baseline.

\subsubsection{Evaluation dataset}

\label{sec:dataset}

\begin{figure}[H]
\begin{center}
\begin{tabular}{c}
\includegraphics[width=0.5\textwidth]{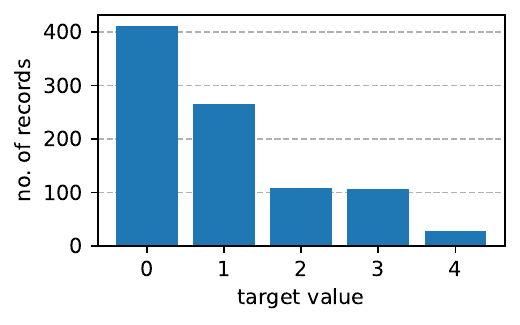}
\end{tabular}
\end{center}
\caption{Distribution of target values, from 0 to 4, where a value of 0 indicates that the patient has no heart disease.}
\label{fig:target_hist}
\end{figure}

The dataset used to test the proposed tool is the UCI Heart Disease Dataset (hereinafter referred to as UCI-HDD) \cite{misc_heart_disease_45}. The target (attribute to be predicted) refers to the presence of heart disease, being an integer ranging from 0 (no presence) to 4. The distribution of target values is displayed in Figure \ref{fig:target_hist}. The dataset has 920 entries of patients from Cleveland, Hungary, VA Long Beach, and Switzerland and is composed of 13 predictive features:

\begin{itemize}
\itemsep0em
\item \textit{age}.
\item \textit{sex}.
\item \textit{cp}, chest pain type.
\item \textit{trestbps}, resting blood pressure on admission (mmHg).
\item \textit{chol}, serum cholesterol (mg/dl).
\item \textit{fbs}, fasting blood sugar is above 120mg/dl (yes, no).
\item \textit{restecg}, resting electrocardiographic results.
\item \textit{thalch}, maximum heart rate achieved.
\item \textit{exang}, exercise-induced angina (yes, no).
\item \textit{oldpeak}, ST depression induced by exercise relative to rest.
\item \textit{slope}, slope of the peak exercise ST segment.
\item \textit{ca}, number of major vessels colored by fluoroscopy.
\item \textit{thal}, thalassemia (normal, fixed or reversible defect).
\end{itemize}

\begin{figure}[H]
\begin{center}
\begin{tabular}{c}
\includegraphics[width=0.8\textwidth]{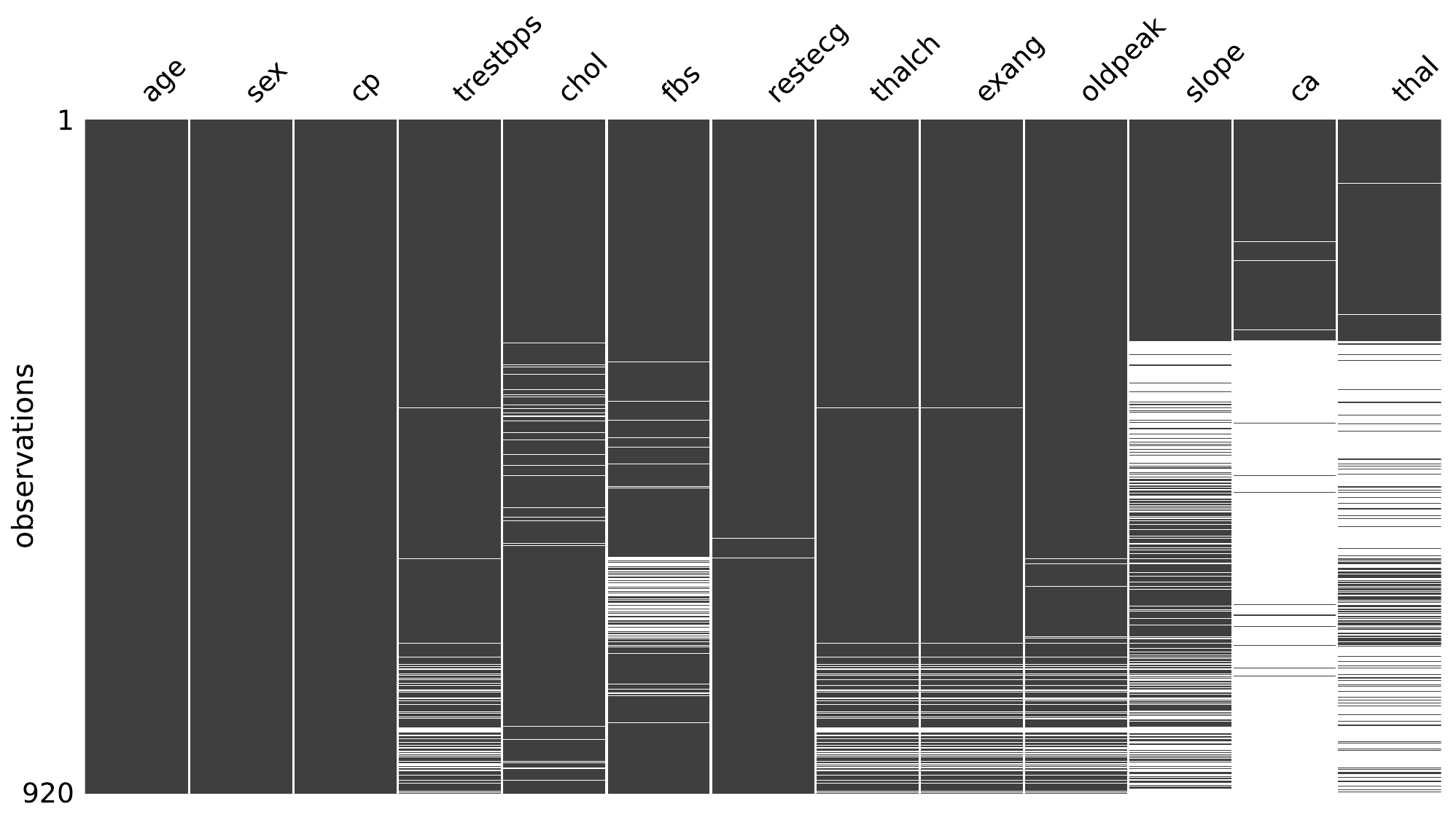}
\end{tabular}
\end{center}
\caption{Matrix distribution of missing (blank) and observed (gray) values across the UCI Heart Disease dataset.}
\label{fig:msno_matrix}
\end{figure}

The fraction of missingness across the whole dataset is approximately 15\%. However, missingness is visibly not evenly distributed, as it can be seen in Figure \ref{fig:msno_matrix}. While some features are totally observed, some few others amass the majority of missing values, as can be checked in Table \ref{tab:feature_missingness}, listing the fraction of missing values by column. Moreover, missing values in these features are not arbitrarily distributed. On the contrary, they appear to share a multivariate connected pattern of missingness.

\begin{table}[h]
	\centering
	\caption{Percentage of missing values by predictive feature in the UCI Heart Disease dataset.}
	\begin{tabular}{lr}
\toprule
Feature &   Missingness (\%) \\
\midrule
age      &   0.0 \\
sex      &   0.0 \\
cp       &   0.0 \\
trestbps &   6.4 \\
chol     &   3.3 \\
fbs      &   9.8 \\
restecg  &   0.2 \\
thalch   &   6.0 \\
exang    &   6.0 \\
oldpeak  &   6.7 \\
slope    &  33.6 \\
ca       &  66.4 \\
thal     &  52.8 \\
\bottomrule
\end{tabular}
	\label{tab:feature_missingness}
\end{table}

To keep the problem simple and the target balanced, we will be using this data to pose and resolve a classification problem. The target value 0 (no presence of heart disease) will be the negative class, and the remaining target values (1 to 4) will be all considered as positive class.

\subsection{Main contributions of the proposed toolbox}

This section summarizes the main list of contributions included in ITI-IQA.
\begin{itemize}
\item The IQA algorithm, which lets us evaluate a set of imputers, both univariate and multivariate, and selects the best strategy for each variable. This selection is made based on multiple criteria, mainly grounded in the similarity between imputed and observed values and bias avoidance. If a minimum data quality threshold is provided, IQA also discards any feature that would not meet this criterion.
\item A new APPRandom imputer, designed to avoid bias as much as possible by filling in missing values with random values while preserving the distribution of the originally observed data.
\item Automatic pipeline construction for dealing with missing data based on the findings made by the IQA algorithm.
\item Flexible configuration files that permit an easy implementation and customization of various own and third-party imputation methods for their evaluation inside IQA.
\item Data quality diagnosing techniques to assess imputed data.
\item A function to estimate the optimum number of iterations for multivariate imputation based on missingness and efficiency.
\end{itemize}

A toolbox containing these utilities and more is expected to be released in the future as a standalone installable Python library, with minimal dependencies and compatible with methods from other widely adopted libraries such as Scipy or Scikit-learn.

\section{Results}

\subsection{IQA results}

\begin{figure}[H]
\begin{center}
\begin{tabular}{c}
\includegraphics[width=0.6\textwidth]{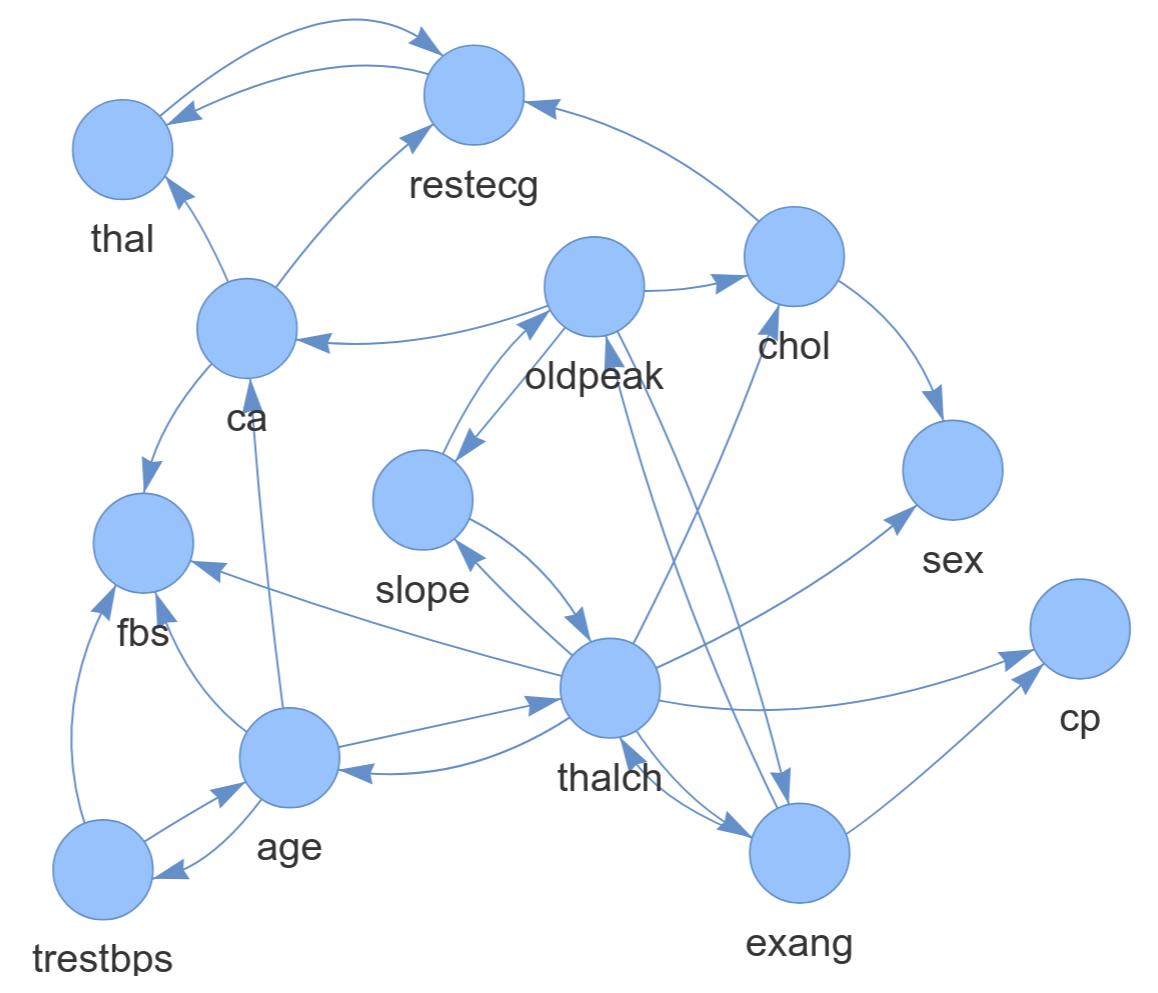}
\end{tabular}
\end{center}
\caption{Graph displaying found feature relations in UCI-HDD that will be acknowledged during multivariate imputation. An edge pointing from A to B indicates that A can be used to estimate the values of B.}
\label{fig:ucihd_relatedvars}
\end{figure}

First, we constructed a dependency graph to discover which features are likely to be accurately inferred from others, as explained in Section \ref{sec:graphs}. Figure \ref{fig:ucihd_relatedvars} shows the dependency graph obtained with UCI-HDD. For each feature, the other features with edges pointing inwards (directly or indirectly through other nodes) will be used in multivariate imputations.

To evaluate the imputation score of each feature ($\delta_x$), a scorer must be chosen. Since many of the features are categorical, we find the $R^2$ score not to be ideal for this case. Instead, we will use a score based on Root Mean Squared Error (RMSE). Since RMSE depends on the range of values of the evaluated vector, we must apply a normalization function so that the results are comparable across different features. Since some features may have an average value close to zero, a typical mean normalization (also known as Scatter index) could lead to unsuitable operations. To avoid this, we normalize RMSE by the difference between maximum and minimum observed values (Equation \ref{eq:nrmse}). Then, we substract the Normalized RMSE (NRMSE) to 1, so that a minimum RMSE of zero would achieve a maximum score of 1.
\begin{equation}
\text{NRMSE}\left(y, \hat{y}\right) = \frac{\text{RMSE}\left(y, \hat{y}\right)}{y_{max} - y_{min}}
\label{eq:nrmse}
\end{equation}

\begin{figure}[H]
\begin{center}
\begin{tabular}{c}
\includegraphics[width=0.75\textwidth]{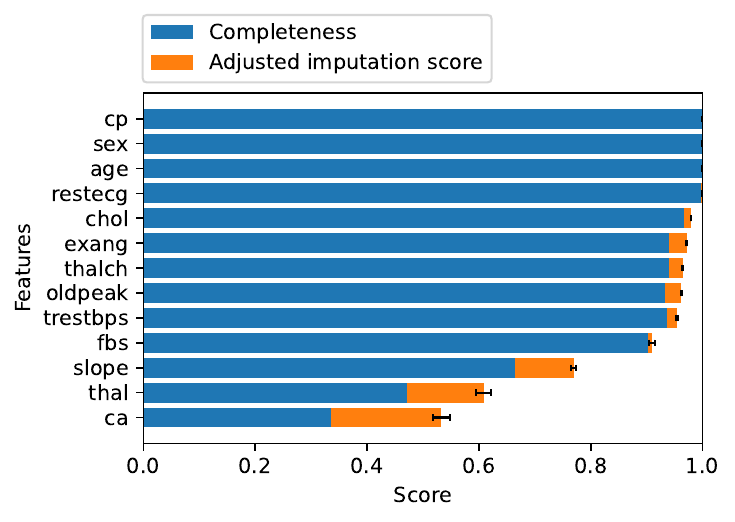}
\end{tabular}
\end{center}
\caption{IQA resulting quality score of UCI-HDD's features.}
\label{fig:ucihd_qualscores}
\end{figure}

After running IQA, the total quality score for each feature (obtained form its completeness and imputation score, as stated in Section \ref{sec:matnmet_alg}) is calculated. Figure \ref{fig:ucihd_qualscores} shows the quality scores for UCI-HDD's features. Table \ref{tab:selected_imputers} shows the selected imputer method for each feature, following the IDs defined in Table \ref{tab:imputers}.

\begin{table}[h]
	\centering
	\caption{IDs of selected imputers for each feature by IQA.}
	\begin{tabular}{lr}
\toprule
Feature &   Imputer ID \\
\midrule
age      &   iter\_br \\
sex      &   iter\_xgb \\
cp       &   iter\_br \\
trestbps &   iter\_br \\
chol     &   iter\_rf \\
fbs      &   iter\_xgb \\
restecg  &   iter\_rf \\
thalch   &   iter\_rf \\
exang    &   iter\_rf \\
oldpeak  &   iter\_rf \\
slope    &  iter\_rf \\
ca       &  iter\_br \\
thal     &  mean \\
\bottomrule
\end{tabular}
	\label{tab:selected_imputers}
\end{table}

\subsection{Comparison with other methods}

\begin{figure}[H]
\begin{center}
\begin{tabular}{c}
\includegraphics[width=0.75\textwidth]{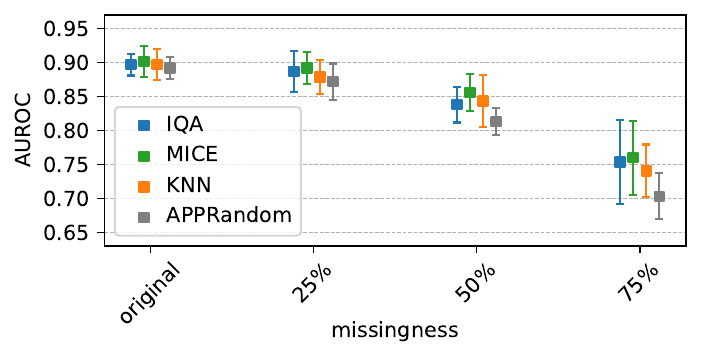}
\end{tabular}
\end{center}
\caption{Mean and 95\% CI of 5-Fold CV AUROC scores of an XGBoost classifier trained and evaluated with imputed data using IQA, MICE, and KNN methods.}
\label{fig:ucihd_aucscores}
\end{figure}

Once the optimal imputation steps recommended by IQA are clear, we can use this set of steps to construct a whole pipeline to deal with missing data. Then, two other pipelines, one using MICE and the other using KNN, are constructed. It is worth mentioning that all three pipelines resort to performing Label Encoding over categorical data so that imputation models can work with them. Figure \ref{fig:ucihd_aucscores} shows the mean AUROC score and 95\% confidence interval of the mean AUROC when training a model to predict the original target (heart disease) using imputed data with each pipeline in a 5-Fold Cross Validation.

Lastly, bias in imputation is assessed using the target substitution method described in Section \ref{sec:evalmethods}, where we try to predict if each entry in each feature of a dataset with filled-in missing values is observed or imputed. Tables \ref{tab:bias_original}, \ref{tab:bias_25}, \ref{tab:bias_50} and \ref{tab:bias_75} show AUROC performances for each feature and imputation strategy for original, 25\%, 50\% and 75\% missingness levels respectively. The average AUROC for each strategy is also presented in the last row of each table. It is worth remembering that a higher AUROC here implies that the imputed values are more easily detected by a capable model, and therefore a lower AUROC metric is desired.

\begin{table}[h!]
	\centering
	\caption{5-Fold mean $\pm$ 95\% CI AUROC when predicting if values are observed or imputed when imputing with each tested strategy at UCI-HDD's original \% of missingness. Features with no or insufficient missing values are left blank.}
	\small
\begin{tabular}{lcccc}
\toprule
{} &          IQA &         MICE &          KNN & APPRandom\\
\midrule
age      &        --- &        --- &        --- &        --- \\
sex      &        --- &        --- &        --- &        --- \\
cp       &        --- &        --- &        --- &        --- \\
trestbps &  0.97$\pm$0.03 &  0.98$\pm$0.04 &  0.98$\pm$0.04 &  0.92$\pm$0.04 \\
chol     &  0.73$\pm$0.22 &  0.85$\pm$0.17 &   0.7$\pm$0.14 &  0.72$\pm$0.12 \\
fbs      &   0.9$\pm$0.04 &        1.0 &  0.97$\pm$0.01 &   0.9$\pm$0.04 \\
restecg  &        --- &        --- &        --- &        --- \\
thalch   &  0.98$\pm$0.03 &   1.0$\pm$0.01 &  0.99$\pm$0.03 &  0.95$\pm$0.04 \\
exang    &  0.98$\pm$0.03 &   1.0$\pm$0.01 &  0.99$\pm$0.03 &  0.95$\pm$0.04 \\
oldpeak  &  0.96$\pm$0.05 &   1.0$\pm$0.01 &  0.97$\pm$0.05 &   0.9$\pm$0.08 \\
slope    &  0.95$\pm$0.02 &        1.0 &  0.99$\pm$0.01 &  0.96$\pm$0.03 \\
ca       &  0.95$\pm$0.02 &        1.0 &       0.99 &  0.94$\pm$0.01 \\
thal     &  0.96$\pm$0.01 &        1.0 &  0.98$\pm$0.01 &  0.88$\pm$0.03 \\
\textbf{average} & \textbf{0.93} & \textbf{0.98} &  \textbf{0.95} &  \textbf{0.90} \\
\bottomrule
\end{tabular}
\label{tab:bias_original}
\end{table}

\begin{table}[h!]
	\centering
	\caption{5-Fold mean $\pm$ 95\% CI AUROC when predicting if values are observed or imputed when imputing with each tested strategy at 25\% level of missingness.}
	\small
\begin{tabular}{lcccc}
\toprule
{} &          IQA &         MICE &          KNN & APPRandom\\
\midrule
age      &  0.98$\pm$0.02 &  0.99$\pm$0.01 &  0.94$\pm$0.03 &  0.59$\pm$0.09 \\
sex      &  0.57$\pm$0.07 &        1.0 &  0.67$\pm$0.04 &  0.59$\pm$0.07 \\
cp       &  0.79$\pm$0.07 &        1.0 &  0.96$\pm$0.04 &  0.63$\pm$0.09 \\
trestbps &  0.97$\pm$0.04 &  0.97$\pm$0.03 &  0.92$\pm$0.05 &  0.69$\pm$0.08 \\
chol     &  0.69$\pm$0.02 &  0.68$\pm$0.04 &  0.74$\pm$0.03 &  0.57$\pm$0.04 \\
fbs      &  0.63$\pm$0.08 &        1.0 &  0.97$\pm$0.03 &  0.63$\pm$0.05 \\
restecg  &        1.0 &        1.0 &  0.89$\pm$0.04 &  0.56$\pm$0.06 \\
thalch   &  0.84$\pm$0.03 &  0.86$\pm$0.05 &   0.9$\pm$0.04 &  0.72$\pm$0.06 \\
exang    &  0.68$\pm$0.07 &        1.0 &  0.94$\pm$0.04 &  0.76$\pm$0.01 \\
oldpeak  &  0.97$\pm$0.02 &  0.97$\pm$0.02 &  0.93$\pm$0.02 &  0.69$\pm$0.02 \\
slope    &  0.82$\pm$0.04 &        1.0 &  0.97$\pm$0.02 &  0.86$\pm$0.02 \\
ca       &   0.9$\pm$0.04 &        1.0 &  0.98$\pm$0.01 &  0.88$\pm$0.04 \\
thal     &  0.93$\pm$0.04 &        1.0 &  0.98$\pm$0.01 &  0.81$\pm$0.04 \\
\textbf{average} & \textbf{0.83} & \textbf{0.96} &  \textbf{0.91} &  \textbf{0.69} \\
\bottomrule
\end{tabular}
\label{tab:bias_25}
\end{table}

\begin{table}[h!]
	\centering
	\caption{5-Fold mean $\pm$ 95\% CI AUROC when predicting if values are observed or imputed when imputing with each tested strategy at 50\% level of missingness.}
	\small
\begin{tabular}{lcccc}
\toprule
{} &          IQA &         MICE &          KNN & APPRandom\\
\midrule
age      &  0.99$\pm$0.01 &  0.98$\pm$0.02 &  0.91$\pm$0.02 &  0.79$\pm$0.08 \\
sex      &  0.53$\pm$0.04 &        1.0 &  0.87$\pm$0.04 &  0.61$\pm$0.05 \\
cp       &  0.69$\pm$0.04 &        1.0 &  0.96$\pm$0.02 &   0.7$\pm$0.03 \\
trestbps &  0.94$\pm$0.02 &  0.94$\pm$0.03 &  0.93$\pm$0.01 &  0.77$\pm$0.06 \\
chol     &  0.68$\pm$0.04 &  0.71$\pm$0.04 &  0.76$\pm$0.02 &   0.7$\pm$0.05 \\
fbs      &  0.55$\pm$0.04 &        1.0 &  0.91$\pm$0.03 &  0.62$\pm$0.07 \\
restecg  &        1.0 &        1.0 &  0.91$\pm$0.04 &  0.73$\pm$0.03 \\
thalch   &  0.84$\pm$0.03 &  0.81$\pm$0.03 &  0.86$\pm$0.03 &  0.82$\pm$0.02 \\
exang    &  0.56$\pm$0.05 &        1.0 &  0.93$\pm$0.02 &  0.66$\pm$0.05 \\
oldpeak  &  0.96$\pm$0.03 &  0.96$\pm$0.03 &  0.92$\pm$0.01 &  0.73$\pm$0.03 \\
slope    &  0.69$\pm$0.06 &        1.0 &  0.94$\pm$0.03 &  0.76$\pm$0.06 \\
ca       &  0.81$\pm$0.05 &        1.0 &  0.95$\pm$0.02 &  0.79$\pm$0.04 \\
thal     &  0.88$\pm$0.04 &        1.0 &  0.95$\pm$0.03 &  0.69$\pm$0.05 \\
\textbf{average} & \textbf{0.78} & \textbf{0.95} &  \textbf{0.91} &  \textbf{0.72}\\
\bottomrule
\end{tabular}
\label{tab:bias_50}
\end{table}

\begin{table}[h!]
	\centering
	\caption{5-Fold mean $\pm$ 95\% CI AUROC when predicting if values are observed or imputed when imputing with each tested strategy at 75\% level of missingness.}
	\small
\begin{tabular}{lcccc}
\toprule
{} &          IQA &         MICE &          KNN & APPRandom\\
\midrule
age      &  0.98$\pm$0.02 &  0.97$\pm$0.02 &  0.91$\pm$0.04 &  0.86$\pm$0.04 \\
sex      &  0.52$\pm$0.01 &        1.0 &  0.91$\pm$0.01 &  0.68$\pm$0.06 \\
cp       &  0.88$\pm$0.03 &        1.0 &  0.95$\pm$0.01 &  0.82$\pm$0.04 \\
trestbps &  0.92$\pm$0.06 &  0.92$\pm$0.05 &  0.95$\pm$0.02 &  0.88$\pm$0.04 \\
chol     &   0.7$\pm$0.03 &  0.74$\pm$0.06 &  0.76$\pm$0.07 &  0.87$\pm$0.05 \\
fbs      &  0.54$\pm$0.05 &        1.0 &  0.83$\pm$0.01 &  0.72$\pm$0.02 \\
restecg  &  0.99$\pm$0.01 &        1.0 &  0.93$\pm$0.02 &  0.86$\pm$0.02 \\
thalch   &  0.76$\pm$0.04 &  0.82$\pm$0.03 &   0.8$\pm$0.04 &  0.89$\pm$0.05 \\
exang    &  0.64$\pm$0.07 &        1.0 &  0.96$\pm$0.01 &  0.74$\pm$0.05 \\
oldpeak  &   0.9$\pm$0.03 &  0.97$\pm$0.02 &  0.91$\pm$0.03 &  0.82$\pm$0.03 \\
slope    &   0.6$\pm$0.06 &        1.0 &  0.94$\pm$0.01 &  0.75$\pm$0.05 \\
ca       &  0.84$\pm$0.02 &        1.0 &  0.93$\pm$0.03 &   0.73$\pm$0.1 \\
thal     &  0.85$\pm$0.06 &        1.0 &  0.95$\pm$0.03 &  0.77$\pm$0.04 \\
\textbf{average} & \textbf{0.78} & \textbf{0.96} &  \textbf{0.90} &  \textbf{0.80}\\
\bottomrule
\end{tabular}
\label{tab:bias_75}
\end{table}

\section{Discussion}
\label{sec:discuss}

The constructed graph of feature dependencies for UCI-HDD is connected, i.e., there are no disconnected ``clusters'' of features. This could make this feature of IQA less notorious since the sets of predecessors of many features are relatively big and close to the complete set of features. However, there are still some features that only have incoming edges, such as \textit{sex} and \textit{fbs}. These features, known as sink features, which do not have any edges leading out to other nodes, will not participate in imputing other features.

In Table \ref{tab:selected_imputers}, we see that multivariate imputation has been found as optimal for most features, with \textit{thal} being the only feature that has been assigned an univariate imputer. It could seem strange that \textit{thal}, which is a categorical feature, is imputed with a simple mean imputer that could produce values outside its encoded labels. However, it is worth bringing back the fact that all results are pseudo-rounded when the data type requires it, like in this case.

Observing Figure \ref{fig:ucihd_aucscores}, one can find that MICE has achieved the best AUROC in all scenarios, although by a small margin, while IQA and KNN alternate in the second spot. These three methods surpass APPRandom, which is depicted here as a more naïve technique, by a noticeable margin. MICE applies multivariate imputation to all features with the whole dataset at its disposal, which seems to gain the upper edge over the masking effect of dependency graphs in IQA. However, maximizing predictive scores is not the only goal (nor the main one) of IQA; it is to cautiously deal with missing data and minimize biases while at it, which leads us to the following results.

After reviewing tables \ref{tab:bias_original} to \ref{tab:bias_75}, we see that IQA achieved a lower average AUROC than MICE and KNN in all four cases when attempting to predict where the missing gaps were. This is a positive result for IQA, since it shows that imputations made by their proposed pipelines are the most indistinguishable from other originally observed values. IQA even manages to score a lower mean AUROC than the APPRandom baseline in the 75\% missingness scenario. While APPRandom is almost ensured to introduce no biases in single features, it can still introduce harmful, recognisable effects multivariate-wise. Since imputations made by APPRandom are random, they do not reproduce correlations nor any kind of multivariate patterns that could be present in the observed data, which a model could recognise. This is why, while serving as a baseline that is easy to understand, it can indeed be surpassed in this aspect. Regarding the MICE strategy, while it always achieved the best predictive AUROC in the previous experiment, it also appears to be the most prone to include biases in posterior analysis when applied over the whole dataset without checking or considering missingness patterns, which could explain its loss in this comparison.

\section{Conclusions}
\label{sec:conclusions}

ITI-IQA is a new set of tools and algorithms for dealing with missing data. Its main focus is not only to estimate values that could be plausible given other real observations but also to do it with caution. The straightforward and thoughtless application of imputation strategies, only trying to minimize errors or maximize the similarity between observed and missing values, can lead to the introduction of serious biases that could, in turn, spoil all subsequent analyses and conclusions.

IQA is highly flexible and customizable, offering various tools for optimizing the design of pipelines for processing data. Moreover, IQA comes with handy visualization tools that enhance its use as an assessment tool. It can be used to evaluate the quality of features using a new quality score based on both missingness and imputation scores. Normally, researchers could be driven to dispose of all features which have a great percentage of missing values. IQA can indicate that some of those features could be saved without the risk of damaging posterior analyses if an imputer has been demonstrated to achieve a higher score than a random imputer while introducing no biases. Using this combined quality score as guidance for including or excluding variables from a study could help researchers preserve most information while still taking minimal risks.

In this study, IQA, MICE, and KNN have been utilized to deal with missing data in the UCI Heart Disease Dataset. While data imputed by IQA has not achieved the best scores when training and evaluating machine learning models with it, IQA's imputation did prove to be the most discrete, as machine learning models had it the most difficult to discern imputed from observed values in the case of IQA. These results align with our leading goal, which is to enable the use of datasets with missing data, not necessarily enhancing it but doing as little damage as possible.

\section{Future work}
\label{sec:future}

Our work exhibits several limitations that we acknowledge in this section while also discussing possible ways of improvement.

Differently handling categorical features poses a difficulty that has been avoided by encoding this data type as numerical. However, the encoding methods often assume a structure that does not hold true, like ordinality. Additionally, handling one-hot-encoded variables could produce unrealistic results (e.g., estimating a positive value for several features, breaking the ``one-hot'' property). Therefore, we acknowledge that handling categorical features is a task that still leaves room for improvement.

In \cite{rubin2018multiple}, stating that ``imputation is not prediction'', the author describes how minimizing an error metric is not a proper solution for finding the best imputer. While our work is grounded in a similar approach, it is also true that we try to avoid biases in other ways. However, it is possible that our methods for handling potential problems arising from biases are not exhaustive enough for some scenarios, and more complex checks could be performed.

In this study, no assumptions have been made about missingness mechanisms, but probably not enough adjustments accounting for them have been implemented either. Many imputation methods (such as multiple imputation) rely on the restrictive assumption of MAR, which often does not hold. When the missingness relates to unobserved data (MNAR), these methods can lead to biased results. In \cite{graham2009missing}, authors argue that the main mechanisms of missingness (MCAR, MAR, and MNAR) should be treated as a continuum between MAR and MNAR, where pure forms of the three mechanisms are never found. Then, the important question to be made is if the violation of the assumptions is big enough to matter.

In some works, an emphasis is made on the hidden information that is present even in missingness and should be accounted for, for it to not be lost in a careless conventional imputation \cite{beesley2021accounting}. Others even hypothesize that preserving the information about missingness as separate attributes (called missing-indicators), for some of it to transcend the imputation phase, can often help achieve greater model performances \cite{lenz2022no}. Future iterations of IQA should accommodate some of the aforementioned considerations in order to deal with missing data appropriately in much more flexible scenarios.

\section*{Acknowledgements}

This work was funded by Generalitat Valenciana (project GVA2024, AI line of research) under grant IMAMCA/2024/11.

We would also like to extend our gratitude to researchers and professors at Universitat Politècnica de València for their valuable help in inspiring and conceiving this work.

\bibliography{main}

\end{document}